\documentclass[10pt,twocolumn]{ICCAS}
\usepackage{diagbox}

\usepackage{makecell}
\usepackage[font=footnotesize]{caption}
\usepackage{subfig}
\usepackage{cite}
\usepackage{array,tabularx}
\usepackage{multicol}
\usepackage{algorithm,algpseudocode}



\begin{document}

\title{Ground-Aware Octree-A* Hybrid Path Planning for Memory-Efficient 3D Navigation of Ground Vehicles}

\author{Byeong-Il Ham${}^{1}$, Hyun-Bin Kim${}^{2}$ and Kyung-Soo Kim${}^{2*}$ }

\affils{ ${}^{1}$Robotics Program, Korea Advanced Institute of Science and Technology(KAIST), \\
Daejeon, 34141, Korea (byeongil\_ham@kaist.ac.kr) \\
${}^{2}$Department of Mechanical Engineering, KAIST, Daejeon, 34141, Korea \\
 (youfree22@kaist.ac.kr and kyungsookim@kaist.ac.kr) {\small${}^{*}$ Corresponding author}}

\abstract{
    In this paper, we propose a 3D path planning method that integrates the A* algorithm with the octree structure. Unmanned Ground Vehicles (UGVs) and legged robots have been extensively studied, enabling locomotion across a variety of terrains. Advances in mobility have enabled obstacles to be regarded not only as hindrances to be avoided, but also as navigational aids when beneficial. A modified 3D A* algorithm generates an optimal path by leveraging obstacles during the planning process. By incorporating a height-based penalty into the cost function, the algorithm enables the use of traversable obstacles to aid locomotion while avoiding those that are impassable, resulting in more efficient and realistic path generation. The octree-based 3D grid map achieves compression by merging high-resolution nodes into larger blocks, especially in obstacle-free or sparsely populated areas. This reduces the number of nodes explored by the A* algorithm, thereby improving computational efficiency and memory usage, and supporting real-time path planning in practical environments. Benchmark results demonstrate that the use of octree structure ensures an optimal path while significantly reducing memory usage and computation time.
}

\keywords{
    A* Algorithm, Octree, Path Planning.
}

\maketitle

%-----------------------------------------------------------------------

\section{Introduction}
Research on the design of Unmanned Ground Vehicles (UGVs)\cite{park2012optimal, tang2023structure} and the locomotion capabilities of legged robots \cite{kim2019highly,grandia2023perceptive,kang2024external} has demonstrated significant advantages over conventional wheeled platforms\cite{biswal2021development}. Rather than merely avoiding terrain obstacles, these robots are capable of directly traversing them. Typical examples include locomotion within confined spaces such as pipes \cite{kim2024development}, as well as overcoming complex obstacles such as high structures, low barriers, and narrow passages \cite{zhuang2023robot}. The maneuverability of these robots enables reconnaissance and rescue operations in environments that are challenging for human access, such as collapsed buildings and mountainous terrain.

Efficient path planning is essential for robots to operate effectively in diverse environments. Despite significant advancements in robotic hardware and control methods, limited battery capacity continues to constrain the operational time of robots. Therefore, generating an optimal path is essential for minimizing travel time. This, in turn, increases the distance that the robot can reach within the limited battery capacity.

Due to limitations in size and power consumption, mobile robots must operate with low computational complexity and memory usage to be applicable in the real-world. The processing of large-scale data for terrain analysis and optimal path computation often requires high-performance hardware, which poses challenges for deployment on systems with limited computational resources.

To address these challenges, we propose a hybrid path planning method that combines the A* algorithm with an octree-based spatial representation. A* is an algorithm that guarantees optimality under an admissible heuristic, while the octree is a data structure that efficiently represents three-dimensional (3D) space by recursively subdividing it into eight subregions. By integrating these two approaches, the proposed method significantly reduces the search time and memory usage of the A* algorithm. Unlike drones, which can freely move in three-dimensional space, legged robots and UGVs are constrained to remain in contact with the ground or can only hover briefly at low altitudes. As a result, their navigable paths are more restricted compared to aerial vehicles. To account for these characteristics, we introduce a height-based penalty into the A* algorithm's cost function, encouraging ground-level path generation. This results in a path planning method optimized for ground vehicles.

Section~\ref{sRELATEDWORKS} provides an overview of related works. Section~\ref{sPATHPLANNING} describes a grid map compression method using an octree structure and introduces an optimal path planning approach based on A* for ground vehicles that leverages obstacles, culminating in the proposed Octree-A* hybrid method. Section~\ref{sBENCHMARK}e presents a performance comparison between the standard A* algorithm and the proposed hybrid algorithm. Finally, Section~\ref{sCONCLUSION} concludes the study.

\section{Related Works}
\label{sRELATEDWORKS}

Various approaches to path planning have been extensively studied. Dijkstra's algorithm can find the optimal path but suffers from high computational time. To address this limitation, the A* algorithm was introduced to reduce computation time. However, if the heuristic function used in A is not admissible, the optimality of the resulting path cannot be guaranteed\cite{candra2020dijkstra}. An improved A* algorithm, which restricts the search to obstacle intersections along a straight line between the start and goal nodes, reduces the average path length by 3\%. Nevertheless, the computation time increases by a factor of four, limiting its applicability in real-time environments\cite{ju2020path}. While A* guarantees optimality under an admissible heuristic, it incurs high computational costs in high-dimensional and unstructured environments.

Rapidly-exploring Random Tree Star~(RRT*) is a sampling-based algorithm that incrementally builds a search tree by randomly sampling the space. It addresses the limitation of the original RRT algorithm, which does not guarantee an optimal solution, and is suitable for use in high-dimensional environments~\cite{karaman2011sampling, solovey2020revisiting}. Despite its asymptotic optimality as the number of samples increases, RRT* often yields suboptimal paths in practice due to limited computational resources. In addition, its performance degrades in environments with narrow passages, and its non-deterministic nature requires further consideration of reliability~\cite{braun2019comparison,belaid2022narrow}. Recent modifications of RRT*~\cite{huang2024adaptive,tu2024improved} showed improved success rates in narrow passage scenarios, addressing limitations of the original method. However, these approaches did not guarantee complete search success.

Although numerous studies have been conducted on path planning in 3D space, most of them focus on underwater environments~\cite{yan2022three, li2023three} or unmanned aerial vehicles (UAVs)~\cite{kiani2021adapted}. In contrast, robots such as quadrupeds and UGVs typically perform path planning on two-dimensional (2D) surfaces~\cite{zhang2021efficient}. Both 2D and 3D path planning commonly utilize algorithms such as RRT and A*.

In many studies~\cite{xu2025path, liu2024safe}, path planning is performed under the assumption that the terrain is continuous, where the robot must either navigate over traversable terrain or avoid untraversable obstacles, often by jumping. For instance, the study by~\cite{zhe2019path} employs an ADFA* algorithm using a simple grid map, focusing solely on 2D motion. 

Other works~\cite{li2022quadruped, li2020grid} construct grid maps using 3D laser radar data and address the challenge of dynamic obstacles in 2D space. Similarly, studies such as~\cite{zhang2021efficient, liu2020research} demonstrate locomotion involving stepping over obstacles or navigating complex 2D terrain. In~\cite{hoeller2024anymal}, terrain information is captured using a scan-dot representation to enable local path planning.

\begin{algorithm}[!t]
\caption{Coordinate-Driven Leaf Search}
\label{aLeafNode}
\begin{algorithmic}[0]
\Require $(x_i, y_i, z_i)$, \; $\mathcal{N}_{R}$, \; $s$
\Ensure $\mathcal{N}_{L}$, \; $(x, y, z) \in \mathcal{N}_{L}$
    \State $(x,y,z) \gets (x_i, y_i, z_i)$
    \While{\textbf{NOT} isLeaf($\mathcal{N}$)}
        \State $s \gets s / 2$
        \State $\mathcal{i}$$ \gets \textbf{ComputeChildIndex}(x,y,z,s)$
        \State $\mathcal{N} \gets \mathcal{N}_{\text{child}}[i]$
        \State $(x, y, z) \gets (x \bmod s,\ y \bmod s,\ z \bmod s)$
    \EndWhile
    \State $\mathcal{N}_L \gets  \mathcal{N}$ 
\end{algorithmic}
\end{algorithm}

\section{Path Planning}
\label{sPATHPLANNING}

In this Section, we introduce the Octree-A* hybrid path planning method. A uniform grid map is first compressed using an octree structure, and then a cost function of A* is proposed to constrain the path to be generated only over valid surface regions, including traversable obstacles. Finally, the integration of the octree representation with the A* algorithm is presented.

\subsection{Octree Structure}

The octree is primarily a compression algorithm applied to densely sampled surface geometry and has the advantage of being lossless~\cite{schnabel2006octree}. It is a hierarchical tree-based data structure for partitioning 3D space, where each node is recursively subdivided into eight cubic child nodes starting from a root node representing a larger region.

Construction of nodes in the octree follows the procedure described below. First, the grid map is uniformly divided into eight nodes. Each node is then checked for the presence of obstacles. If any obstacle is present within a node, it is further subdivided into eight child nodes. If no obstacle is present—i.e.,the node contains only empty space, start, or goal—it is not further subdivided. A leaf node represents a terminal region in the space that is no longer subdivided. Each newly generated child node is recursively processed until it becomes a leaf node or no obstacles are present within the node. If a child node is identical to a uniform node, it is not further subdivided.

\begin{algorithm}[t!]
\caption{Get Neighbor Leaf Node}
\label{aGetNeighbors}
\begin{algorithmic}[0]
\Require $\mathcal{N}_{L}$, \; $(x, y, z) \in \mathcal{N}_{L}$
\Ensure $\mathcal{O}$
    \State $\mathcal{O} \gets \emptyset$
    \State $ds \gets s/2 \in \mathcal{N}_{L}$
    \State $\mathcal{M}$ $ \gets \textbf{ComputeAdjacents}(x,y,z,ds+r)$
    \For{$m$ $ \in \mathcal{M}$} 
        \State $\mathcal{N}$ $ \gets \textbf{Algorithm~\ref{aLeafNode}}(m)$
        \If {isValid($\mathcal{N}$)}
            % \State $\mathcal{O} \gets \mathcal{N}$
            \State $\mathcal{O} \gets \mathcal{O} \cup \{\mathcal{N}\}$
        \EndIf
    \EndFor

\end{algorithmic}
\end{algorithm}

\subsection{A* Algorithm}

\begin{figure*}[t!]
    \centering
    \subfloat[Scenario 1\label{fAstar1}]{
        \includegraphics[width=0.46\textwidth]{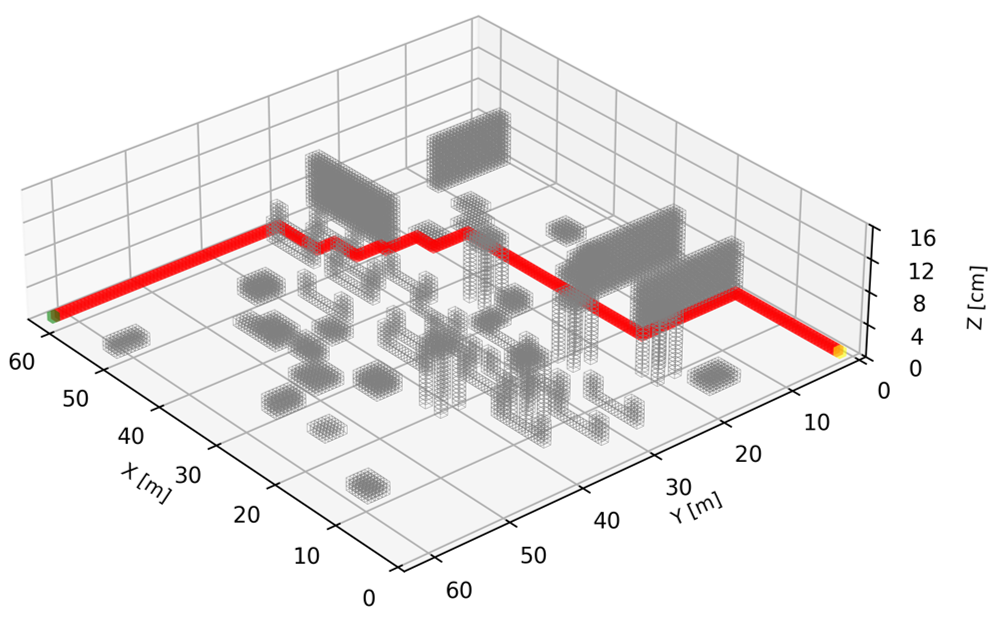}
    }\hfill
    \subfloat[Scenario 2\label{fAstar2}]{
        \includegraphics[width=0.46\textwidth]{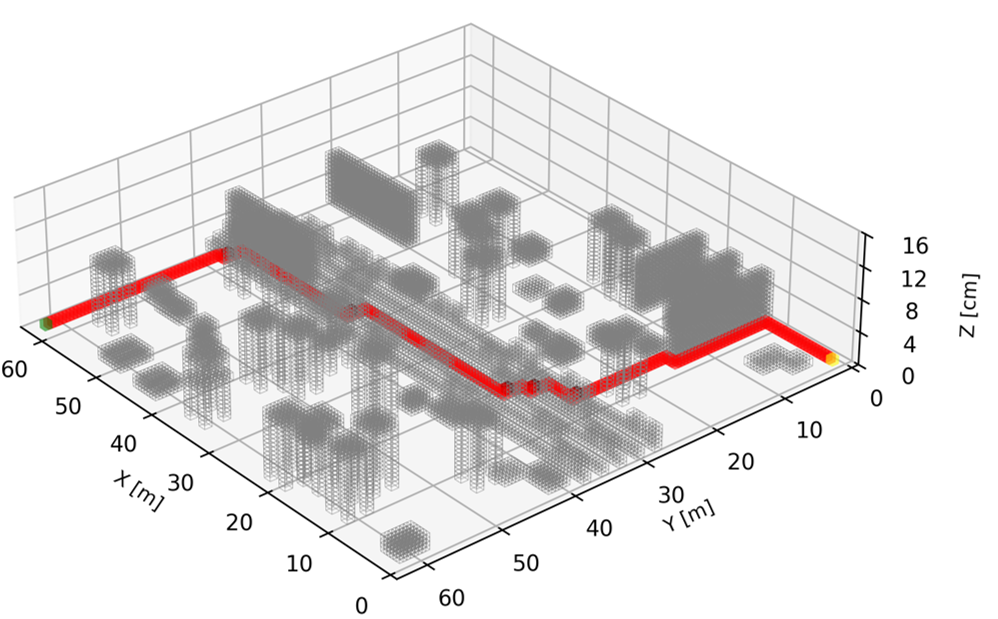}
    }
    \caption{Results of the modified 3D A* algorithm executed on a uniform grid, showing the generated path. Gray represents obstacles, red and black indicate the planned path, yellow denotes the start node, and green denotes the goal node.}
    \label{fAstar}
\end{figure*}

To ensure successful path planning and produce shorter trajectories, A* algorithm was utilized. Unlike UAVs and manipulators, ground vehicles are subject to the constraint of moving on the ground surface. The modified cost function applies the above constraint as a soft constraint within the node exploration process. The cost function is defined as follows:
\begin{equation}
    \min_{n \in N} \quad g(n) + h(n) + \alpha r(n)
    \label{eCostfunction}
\end{equation}
\noindent
where $N$ is the set of neighbor nodes, $n \in N$ is a candidate node, $g(n)$ is the accumulated cost from the start node to node $n$, and $h(n)$ is the heuristic cost defined as the Manhattan distance between node $n$ and the goal node, $\alpha$ is positive scalar penalty weight, and $r(n)$ is a penalty term defined as the vertical distance from the center of node $n$ to the nearest surface. The penalty term is designed to achieve two objectives. First, it increases the cost of nodes that are distant from the surface, discouraging the selection of nodes that are not on the ground. Second, it prevents path generation toward heights that are beyond the robot's traversable capability, effectively discouraging upward movement from the current position when such movement is infeasible.

\subsection{Octree-A* Algorithm}

\begin{table}[!t]
\renewcommand{\arraystretch}{1.2}
\caption{Comparison of A* and Octree-A* performance in two benchmark scenarios.}
\label{tBenchmark}
\begin{tabular}{ccc|ccc}
\Xhline{1.1pt}
            & S1         & S2         & S1            & S2            &      \\
            & \multicolumn{2}{c|}{A*} & \multicolumn{2}{c}{Octree-A*} & Unit \\ \hline
Path length & 121.5      & 125        & 122           & 124           & m    \\
Comp. time  & 927.7      & 7495       & 68.4          & 661           & ms   \\
Memory      & 8192       & 8192      & 367.4          & 608.7         & kB   \\ \Xhline{1.1pt}
\end{tabular}
\end{table}

To integrate the octree with the A*, several challenges must be addressed. In contrast to uniform grids, leaf nodes in an octree have irregular neighbor configurations, where both the number and size of neighboring nodes may differ. Therefore, applying the A* to octree-based grid, it is essential to identify both the neighboring nodes and their corresponding sizes.

An octree node $\mathcal{N}$ consists of a leaf indicator, a state, and a set of child nodes. The state can encode whether the node is empty, occupied, or corresponds to a start or goal. The pseudocode of the method for finding the leaf node $\mathcal{N}_{L}$ that contains a given 3D coordinate $(x_i, y_i, z_i)$ is described in Algorithm~\ref{aLeafNode}. The procedure recursively traverses the octree beginning at the root node $\mathcal{N}_R$ with side length $s$. If the node is not a leaf node, update $s$ to represent the side length of its child nodes. Among the eight child nodes, identify the index $\mathcal{i}$ such that the coordinate $(x, y, z)$ lies within child node $\mathcal{N}_{\text{child}}[i]$, and update node accordingly. Then, update $(x, y, z)$ to the center coordinates within the selected child node using modulo operation. The algorithm returns the leaf node when a leaf node is reached.

Algorithm~\ref{aGetNeighbors} presents the procedure for identifying neighbor leaf nodes of a given leaf node. The set $\mathcal{O}$ is initialized to store valid neighboring leaf nodes, and the variable $ds$ is assigned half the side length of the input leaf node. The ComputeAdjacents function computes the nearest coordinates in each feasible direction using the center coordinate of the current leaf node, $ds$, and the minimum grid resolution $r$. The resulting coordinates are stored in the set $\mathcal{M}$. For each element in $\mathcal{M}$, Algorithm~\ref{aLeafNode} is used to identify the leaf node that contains the corresponding coordinate. Each leaf node is checked using the isValid function, and only those without obstacles are pushed into the $\mathcal{O}$. As a result, $\mathcal{O}$ contains only the coordinates of candidate nodes, which can be directly used in the A*.

\begin{figure*}[t!]
    \centering
    \subfloat[Scenario 1\label{fOctree1}]{
        \includegraphics[width=0.46\textwidth]{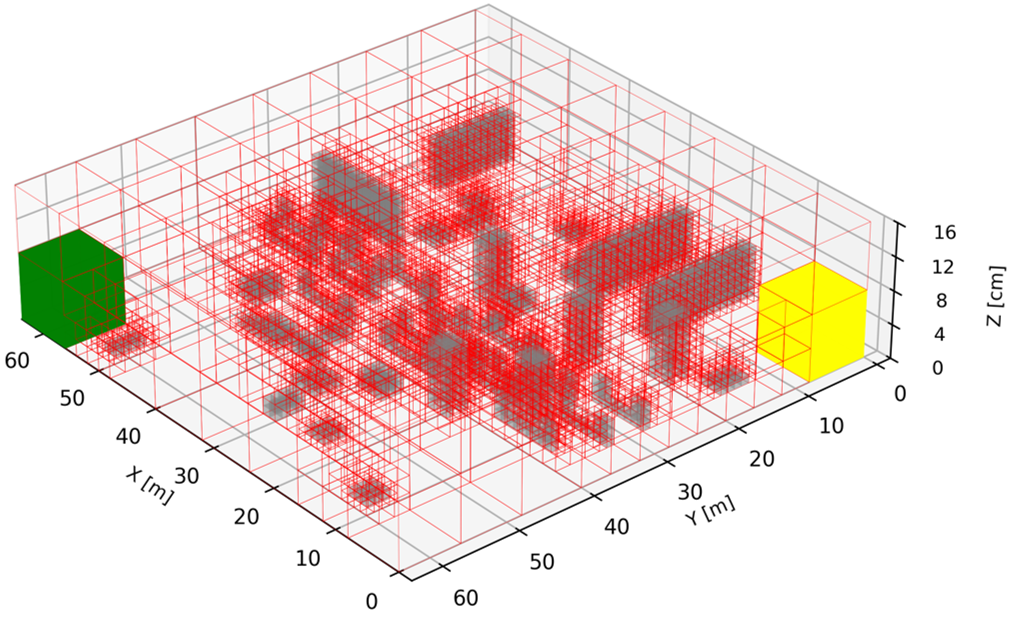}
    }\hfill
    \subfloat[Scenario 2\label{fOctree2}]{
        \includegraphics[width=0.46\textwidth]{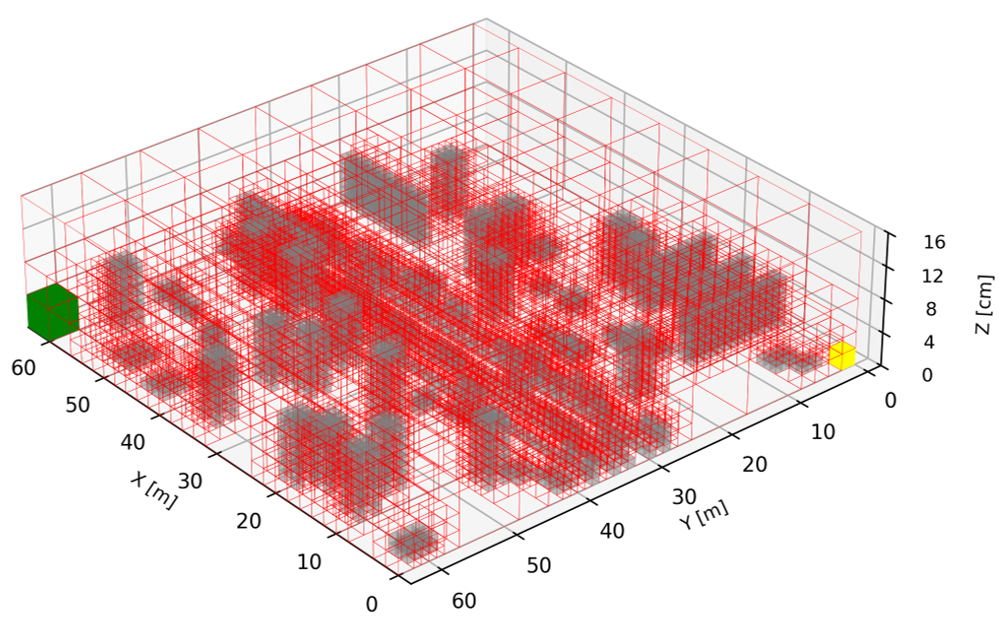}
    }
    \caption{Results of the octree applied to each scenario. Red outlines indicate the edges of leaf nodes, while gray, yellow, and green follow the same color scheme as in Fig.~\ref{fAstar}.}
    \label{fOctree}
\end{figure*}

\begin{figure*}[t!]
    \centering
    \subfloat[Scenario 1\label{scenario1}]{
        \includegraphics[width=0.46\textwidth]{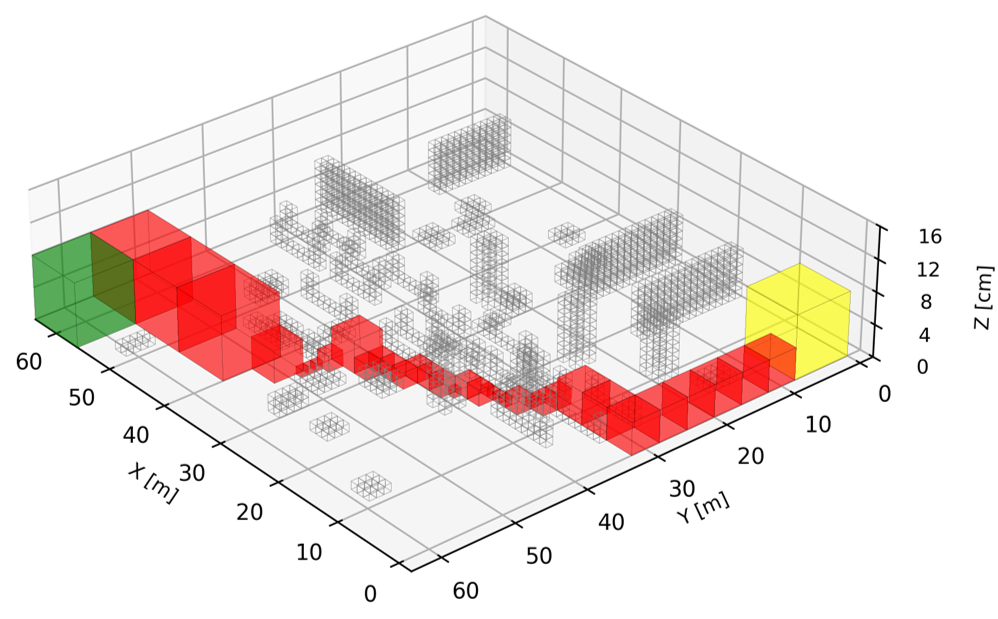}
    }\hfill
    \subfloat[Scenario 2\label{scenario2}]{
        \includegraphics[width=0.46\textwidth]{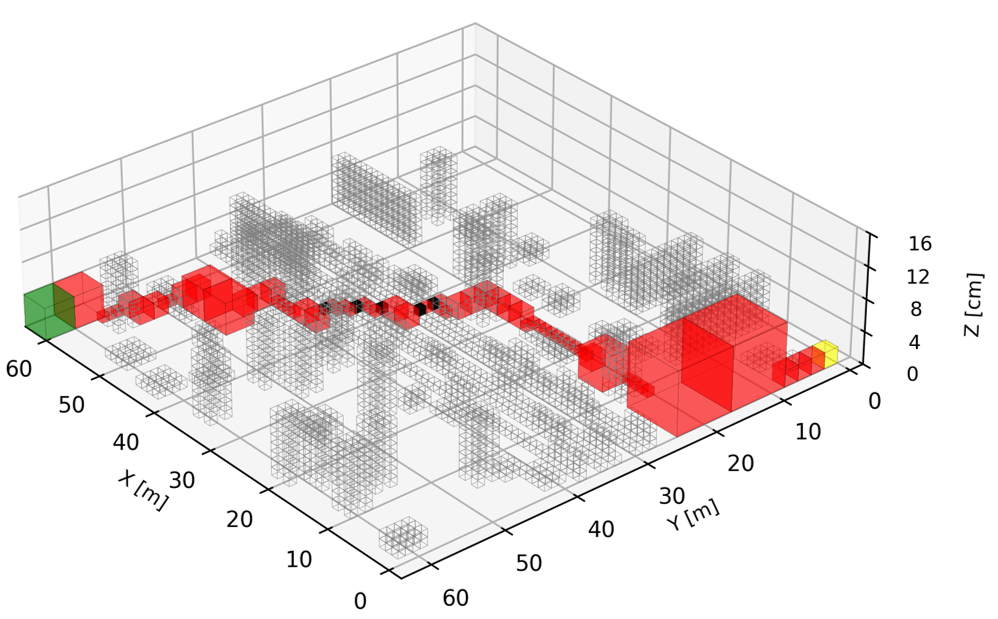}
    }
    \caption{Illustration of the Octree-A*. The color scheme is identical to that used in Fig.~\ref{fAstar}.}
    \label{fOcastar}
\end{figure*}

\section{Benchmark}
\label{sBENCHMARK}
Simulations were conducted on Intel i7-9700 CPU running Ubuntu 22.04 operating system. The algorithms were implemented in C++ and executed using a compiler compliant with the C++17 standard. Two simulation scenarios were conducted. The first scenario represents a condition where an optimal path can be generated without leveraging obstacles. In contrast, the second scenario requires obstacle leveraging—i.e., the path can only be generated by traversing over obstacles. Each uniform grid cell has a size of 0.5m$\times$0.5m$\times$0.5cm, and the map consists of 128$\times$128$\times$128 grid cells. For both A* and Octree-A*, the $g(n)$ is computed by accumulating the distances between the midpoints of adjacent nodes.

Fig.~\ref{fAstar} illustrates the path computed by the modified 3D A* algorithm on a uniform grid. In the path visualization, red segments indicate traversal along the ground surface, while black segments represent portions of the path over obstacle surfaces. In both scenarios, the paths satisfy the constraint of moving along the surface while maintaining optimality. In particular, Fig.~\ref{fAstar2} shows that the path is generated only over obstacles that are feasible to leverage. Fig.~\ref{fOctree} visualizes the size and distribution of leaf nodes for each scenario. Scenario 1 contains 47 obstacles, whereas Scenario 2 contains 78. As shown in the figure, a lower number of obstacles results in fewer leaf nodes with larger average side lengths. The visualization of the modified A* applied to leaf nodes is shown in Fig.~\ref{fOcastar}. Although the generated paths differ from those on the uniform grid, they satisfy all constraints and successfully achieve optimality. Table~\ref{tBenchmark} presents the quantitative benchmark results of A* and Octree-A*.

Compared to uniform nodes, applying leaf nodes resulted in a 0.41\% increase in path length for Scenario 1 and a 0.80\% decrease for Scenario 2 indicating that the path lengths generated by both node types are comparable. The computation time of Octree-A* is the combined time of both octree generation and A* path planning. The Octree-A* significantly reduced computation time by 92.63\% and 91.18\% in Scenarios 1 and 2, respectively. Memory usage was reduced by 95.50\% and 92.73\% in Scenarios 1 and 2, respectively, with higher compression observed in environments containing fewer obstacles.

\section{CONCLUSION}
\label{sCONCLUSION}
In this paper, we studied the 3D optimal path planning method for ground vehicles that generates paths by leveraging obstacles when beneficial. The cost function was modified with a penalty to enforce surface-level path generation and to restrict traversal to only feasible obstacles. With the octree structure, memory consumption dropped to under 10\% while computation time was reduced by a factor greater than 10. The improvements in performance were achieved without compromising the characteristics of A*. Since all path lengths were computed by connecting the center points of nodes, the trajectories may not be shortest for ground vehicles. By applying post-processing such as spline curves, shorter and smoother paths may be obtained.

\end{document}